\documentclass[runningheads]{llncs}

\usepackage{graphicx}
\usepackage{hyperref}
\usepackage{xcolor}
\usepackage{textcomp}
\usepackage{bbding}
\usepackage{amsfonts}
\usepackage{amsmath}
\usepackage{amssymb}
\usepackage{mathtools}
\usepackage{pgf}
\usepackage{tikz}
\usetikzlibrary{arrows, automata, shapes, petri, positioning, calc}
\usepackage[font=scriptsize,labelfont=bf]{caption}
\usepackage{fancyvrb}
\usepackage{cite}

\hypersetup{
	colorlinks,
	linkcolor={red!50!black},
	citecolor={blue!60!black},
	urlcolor={blue!80!black},
	pdfauthor={Pegoraro Marco, Uysal Merih Seran, van der Aalst Wil M.P.},
	pdftitle={PROVED: A Tool for Graph Representation and Analysis of Uncertain Event Data},
	pdfsubject={Process Mining over Uncertain Data},
	pdfkeywords={Process Mining, Uncertain Data, Partial Order, Petri Net Tool},
	pdfproducer={LaTeX},
	pdfcreator={pdfLaTeX}
}

\begin{document}

\title{PROVED: A Tool for Graph Representation and Analysis of Uncertain Event Data\thanks{
		Postprint version.
		In \emph{International Conference on Application and Theory of Petri Nets and Concurrency} (Petri Nets), 2021.
		\textsuperscript{\textcopyright}Springer.
		We thank the Alexander von Humboldt (AvH) Stiftung for supporting our research interactions. Please do not print this document unless strictly necessary.
}
}

\author{Marco Pegoraro~\Envelope\orcidID{0000-0002-8997-7517} \and Merih Seran Uysal\orcidID{0000-0003-1115-6601} \and Wil M.P. van der Aalst\orcidID{0000-0002-0955-6940}}

\authorrunning{Pegoraro et al.}
\titlerunning{PROVED: A Tool for Uncertain Event Data}

\institute{Process and Data Science Group (PADS) \\ Department of Computer Science, RWTH Aachen University, Aachen, Germany
	\email{\{pegoraro,uysal,wvdaalst\}@pads.rwth-aachen.de}\\
	\url{http://www.pads.rwth-aachen.de/}}

\maketitle

\begin{abstract}
The discipline of process mining aims to study processes in a data-driven manner by analyzing historical process executions, often employing Petri nets. Event data, extracted from information systems (e.g. SAP), serve as the starting point for process mining. Recently, novel types of event data have gathered interest among the process mining community, including uncertain event data. Uncertain events, process traces and logs contain attributes that are characterized by quantified imprecisions, e.g., a set of possible attribute values. The PROVED tool helps to explore, navigate and analyze such uncertain event data by abstracting the uncertain information using behavior graphs and nets, which have Petri nets semantics. Based on these constructs, the tool enables discovery and conformance checking.

\end{abstract}

\section{Introduction}
Process mining is a branch of process sciences that performs analysis on processes focusing on a log of execution data.~\cite{van2016process} From an event log of the process, it is possible to automatically discover a model that describes the flow of a case in the process, or measure the deviations between a normative model and the log.

The primary enabler of process mining analyses is the control-flow perspective of event data, which has been extensively investigated and utilized by researchers in this domain. 

Modern information systems supporting processes can enable the extraction of more data perspectives: for instance, it is often possible to retrieve (and thus analyze) additional event attributes, such as the agent (resource) associated with the event, or the cost of a specific activity instance.

Collected event data can be subjected to errors, imprecisions and anomalies; as a consequence, they can be affected by uncertainty. Uncertainty can be caused by many factors, such as sensitivity of sensors, human error, limitations of information systems, or failure of recording systems. The type of uncertainty we consider here is quantified: the event log includes some meta-attributes that describe the uncertainty affecting the event. For instance, the activity label of an event can be unknown, but we might have access to a set of possible activity labels for the event. In this case, in addition to the usual attributes constituting the event in the log, we have a meta-attribute containing a set of activity labels associated with the event. In principle, such meta-attributes can be natively supported by the information system; however, they are usually inferred after the extraction of the event log, in a pre-processing step to be undertaken before the analysis. Often, this pre-processing step necessitates domain knowledge to define, identify, and quantify different types of uncertainty in the event log.

In an event log, regular traces provide a static description of the events that occurred during the completion of a case in the process. Conversely, uncertain process traces contain behavior, and describe a number of possible scenarios that might have occurred in reality. Only one of these scenarios actually took place. It is possible to represent this inherent behavior of uncertain traces with graphical constructs, which are built from the data available in the event log. Some applications of process mining to uncertain data require a model with execution semantics, so to be able to execute all and only the possible real-life scenarios described by the uncertain attributes in the log. To this end, Petri nets are the model of choice to accomplish this, thanks to their ability to compactly represent complex constructs like exclusive choice, possibility of skipping activities, and most importantly, concurrency.

Process mining using uncertain event data is an emerging topic with only a few recent papers. The topic was first introduced in~\cite{pegoraro2019mining} and successively extended in~\cite{pegoraro2020conformance}: here, the authors provide a taxonomy and a classification of the possible types of uncertainty that can appear in event data. Furthermore, they propose an approach to obtain measures for conformance score (upper and lower bounds) between uncertain process traces and a normative process model represented by a Petri net.

An additional application of process mining algorithms for uncertain event logs relates to the domain of process discovery. Here, the uncertain log is mined for possible directly-follows relationships between activities: the result, an Uncertain Directly-Follows Graph (UDFG), expresses the minimum and maximum possible strength of the relationship between pair of activities. In turn, this can be exploited to perform process discovery with established discovery techniques. For instance, the inductive miner algorithm can, given the UDFG and some filtering parameters, automatically discover a process model of the process which also embeds information about the uncertain behavior~\cite{pegoraro2019discovering}.

While the technological sector of process mining software has been flourishing in recent years, no existing tool -- to the best of our knowledge -- can analyze or handle event data with uncertainty. In this paper, we present a novel tool based on Petri nets, which is capable of performing process mining analyses on uncertain event logs. The PROVED (PRocess mining OVer uncErtain Data) software~\cite{proved} is able to leverage uncertain mining techniques to deliver insights on the process without the need of discarding the information affected by uncertainty; on the contrary, uncertainty is exploited to obtain a more precise picture of all the possible behavior of the process. PROVED utilizes Petri nets as means to model uncertain behavior in a trace, associating every possible scenario with a complete firing sequence. This enables the analysis of uncertain event data.

The remainder of the paper is structured as follows: Section~\ref{sec:related} provides an overview of the relevant literature on process mining over uncertainty. Section~\ref{sec:preliminaries} presents the concept of uncertain event data with examples. Section~\ref{sec:architecture} illustrates the architectural structure of the PROVED tool. Section~\ref{sec:usage} demonstrates some uses of the tool. Lastly, Section~\ref{sec:conclusions} concludes the paper.

\section{Related Work}\label{sec:related}
The problem of modeling systems containing or representing uncertain behavior is well-investigated and has many established research results. Systems where specific components are associated with time intervals can, for instance, be modeled with time Petri nets~\cite{berthomieu1991modeling}. Large systems with more complex timed inter-operations between components can be represented by interval-timed coloured Petri nets~\cite{van1993interval}. Probabilistic effects can be modeled and simulated in a system by formalisms such as generalized stochastic Petri nets~\cite{marsan1998modelling}. It is important to notice, however, that the focus of process mining over uncertain event data is different: the aim is not to simulate the uncertain behavior in a model, but rather to perform data-driven analyses, some results of which can be represented by (regular) Petri nets.

The PROVED tool contains the implementation of existing techniques for process mining over uncertain event data. In this paper, we will show the capabilities of PROVED in performing the analysis presented in the literature mentioned above. In terms of tool functionalities, constructing a Petri net based on the description of specific behavior -- known as synthesis in Petri net research -- has some precedents: for instance, from transition systems~\cite{carmona2009genet} in the context of process discovery. More relevantly for this paper, the VipTool~\cite{bergenthum2008synthesis} allows to synthesize Petri nets based on partially ordered objects. While partial order between events is in itself a kind of uncertainty and a consequence of the presence of uncertain timestamps, in this tool paper we extend Petri net synthesis to additional types of uncertainty, and we add process mining functionalities.

\section{Preliminary Concepts}\label{sec:preliminaries}
The motivating problem behind the PROVED tool is the analysis of uncertain event data. Let us give an example of a process instance generating uncertain data.

An elderly patient enrolls in a clinical trial for an experimental treatment against myeloproliferative neoplasms, a class of blood cancers. The enrollment in this trial includes a lab exam and a visit with a specialist; then, the treatment can begin. The lab exam, performed on the 8th of July, finds a low level of platelets in the blood of the patient, a condition known as thrombocytopenia (TP). At the visit, on the 10th of May, the patient self-reports an episode of night sweats on the night of the 5th of July, prior to the lab exam: the medic notes this, but also hypothesized that it might not be a symptom, since it can be caused not by the condition but by external factors (such as very warm weather). The medic also reads the medical records of the patient and sees that, shortly prior to the lab exam, the patient was undergoing a heparine treatment (a blood-thinning medication) to prevent blood clots. The thrombocytopenia found with the lab exam can then be primary (caused by the blood cancer) or secondary (caused by other factors, such as a drug). Finally, the medic finds an enlargement of the spleen in the patient (splenomegaly). It is unclear when this condition has developed: it might have appeared at any moment prior to that point. The medic decides to admit the patient to the clinical trial, starting 12th of July.

These events are collected and recorded in the trace shown in Table~\ref{table:uncertaintrace} in the information system of the hospital. Uncertain activities are indicated as a set of possibilities. Uncertain timestamps are denoted as intervals. Some event are indicated with a ``?'' in the rightmost column; these so-called \emph{indeterminate events} have been recorded, but it is unclear if they actually happened in reality. Regular (i.e., non-indeterminate) events are marked with ``!''. For the sake of readability, the timestamp field only indicates the day of the month.


\begin{table}[]
	\caption{The uncertain trace of an instance of healthcare process used as a running example. For the sake of clarity, we have further simplified the notation in the timestamps column, by showing only the day of the month.}
	\label{table:uncertaintrace}
	\centering
	\begin{tabular}{ccccc}
		\textbf{Case ID}        & \textbf{Event ID} & \textbf{Timestamp}                                                                                                     & \textbf{Activity}             & \multicolumn{1}{l}{\textbf{Indet. event}} \\ \hline
		\multicolumn{1}{|c|}{ID192} & \multicolumn{1}{c|}{$e_1$} 
		& \multicolumn{1}{c|}{5}                                                                         & \multicolumn{1}{c|}{\emph{NightSweats}}        & \multicolumn{1}{c|}{?}                    \\ \hline
		\multicolumn{1}{|c|}{ID192}& \multicolumn{1}{c|}{$e_2$} & \multicolumn{1}{c|}{8}                                                                         & \multicolumn{1}{c|}{\{\emph{PrTP}, \emph{SecTP}\}} & \multicolumn{1}{c|}{!}                    \\ \hline
		\multicolumn{1}{|c|}{ID192}& \multicolumn{1}{c|}{$e_3$} & \multicolumn{1}{c|}{[4, 10]}                                                                         & \multicolumn{1}{c|}{\emph{Splenomeg}} & \multicolumn{1}{c|}{!}                    \\ \hline
		\multicolumn{1}{|c|}{ID192}& \multicolumn{1}{c|}{$e_4$} & \multicolumn{1}{c|}{12}                                                                         & \multicolumn{1}{c|}{\emph{Adm}}        & \multicolumn{1}{c|}{!}                    \\ \hline
	\end{tabular}
\end{table}

Throughout the paper, we will utilize the trace of Table~\ref{table:uncertaintrace} as a running example to showcase the functionalities of the PROVED tool.

\section{Architecture}\label{sec:architecture}
This section provides an overview of the architecture of the PROVED tool, as well as a presentation of the libraries and existing software that are used in the tool as dependencies.

Our tool has two distinct parts, a library (implemented in the PROVED Python package) and a user interface allowing to operate the functions in the library in a graphical, non-programmatic way.

The library is written in the Python programming language (compatible with versions 3.6.x through 3.8.x), and is distributed through the Python package manager \texttt{pip}~\cite{pip}. Notable software dependencies include:
\begin{itemize}
	\item PM4Py~\cite{berti2019process}: a process mining library for Python. PM4Py is able to provide many classical process mining functionalities needed for PROVED, including importing/exporting of logs and models, management of log objects, and conformance checking through alignments. Notice that PM4Py also provides functions to represent and manage Petri nets.
	\item NetworkX~\cite{hagberg2008exploring}: this library provides a set of graph algorithms for Python. It is used for the management of graph objects in PROVED.
	\item Graphviz~\cite{ellson2001graphviz}: this library adds visualization functionalities for graphs to PROVED, and is used to visualize directed graphs and Petri nets.
\end{itemize}
The aforementioned libraries enable the management, analysis and visualization of uncertain event data, and support the mining techniques of the PROVED toolset here illustrated. An uncertain log in PROVED is a log object of the PM4Py library; here, we will list only the novel functionalities introduced in PROVED, while omitting existing features inherited from PM4PY -- such as importing/exporting and attribute manipulation.

	\subsection{Artifacts}\label{sec:artifacts}
	As mentioned earlier, uncertain data contain behavior and, thus, dedicated constructs are necessary to enable process mining analysis. In the PROVED tool, the subpackage \texttt{proved.artifacts} contain the models and construction methods of such constructs. Two fundamental artifacts for uncertain data representation are available:
	\begin{itemize}
		\item \texttt{proved.artifacts.behavior\_graph}: here are collected the PROVED functionalities related to the \emph{behavior graph} of an uncertain trace. Behavior graphs are directed acyclic graphs that capture the variability caused by uncertain timestamps in the trace, and represent the partial order relationships between events. The behavior graph of the trace in Table~\ref{table:uncertaintrace} is shown in Figure~\ref{fig:behgraph}. The PROVED library can build behavior graphs efficiently (in quadratic time with respect to the number of events) by using an algorithm described in~\cite{pegoraro2020efficient}.
		\item \texttt{proved.artifacts.behavior\_net}: this subpackage includes all the functionalities necessary to create and utilize \emph{behavior nets}, which are acyclic Petri nets that can replay all possible sequences of activities (called \emph{realizations}) contained in the uncertain trace. Behavior nets allow to simulate all ``possible worlds'' described by an uncertain trace, and are crucial for tasks such as computing conformance scores between uncertain traces and a normative model. The construction technique for behavior nets is detailed in~\cite{pegoraro2020conformance}.
	\end{itemize}
	
	\begin{figure}
		\centering
		\begin{minipage}[t]{0.48\textwidth}
			\centering
			\begin{tikzpicture}[->, node distance=3.5cm, nodes={draw, ellipse, scale=.65}]
			
			\node[dashed]	(A)	[label=below:$e_1$]								{$\text{NightSweats}$};
			\node			(B)	[right of=A, label=below:$e_2$]					{$\{\text{PrTP, SecTP}\}$};
			\node			(C)	[below of=B, yshift=1.5cm, label=below:$e_3$]	{$\text{Splenomeg}$};
			\node			(D)	[right of=C, yshift=1cm, label=below:$e_4$]		{$\text{Adm}$};
			
			\path
			(A) edge (B)
			(B) edge (D)
			(C) edge (D);
			\end{tikzpicture}
			\caption{The behavior graph of the trace in Table~\ref{table:uncertaintrace}. All the nodes in the graph are connected based on precedence relationships. Pairs of nodes for which the order is certain are connected by a path in the graph; pairs of nodes for which the order is unknown are pairwise unreachable.}
			\label{fig:behgraph}
		\end{minipage}\hfill
		\begin{minipage}[t]{0.48\textwidth}
			\centering
			\begin{tikzpicture}[node distance=.3cm and .35cm, >=stealth', , nodes={scale=.58}]
			
			\tikzstyle{place} = [circle,draw,thick,minimum size=6mm]
			\tikzstyle{transition} = [rectangle,draw,thick,minimum size=4mm]
			\tikzstyle{invisible} = [transition, fill=black]
			\tikzstyle{finaltoken} = [token, fill=black!30]
			
			\node [place,tokens=1] (p1) [] {};
			
			\node [transition] (t1) [above right= of p1, label=above:{$e_1$}] {NightSweats};
			\draw [->] (p1) to (t1.west);
			
			\node [invisible] (t2) [below right= of p1, label=above:{$e_1$}] {NightSweats};
			\draw [->] (p1) to (t2.west);
			
			\node [place] (p2) [below right= of t1] {};
			\draw [->] (t1.east) to (p2);
			\draw [->] (t2.east) to (p2);
			
			\node [transition] (t3) [above right= of p2, label=above:{$e_2$}] {PrTP};
			\draw [->] (p2) to (t3.west);
			
			\node [transition] (t4) [below right= of p2, label=above:{$e_2$}] {SecTP};
			\draw [->] (p2) to (t4.west);
			
			\node [place] (p3) [below right= of t3] {};
			\draw [->] (t3.east) to (p3);
			\draw [->] (t4.east) to (p3);
			
			\node [place,tokens=1] (p4) [below left= of t2] {};
			
			\node [place] (p5) [below right= of t4] {};
			
			\node [transition] (t5) at ($(p4)!0.55!(p5)$) [label=above:{$e_3$}] {Splenomeg};
			\draw [->] (p4) to (t5);
			\draw [->] (t5) to (p5);
			
			\node [transition] (t6) [above right= of p5, label=above:{$e_4$}] {Adm};
			\draw [->] (p3) to (t6.north west);
			\draw [->] (p5) to (t6.south west);
			
			\node [place] (p6) [right= of t6] {};
			\draw [->] (t6) to (p6);
			\node [finaltoken] at (p6) {};
			\end{tikzpicture}
			\caption{The behavior net corresponding to the uncertain trace in Table~\ref{table:uncertaintrace}. The labels above the transitions show the corresponding uncertain event. The initial marking is displayed; the gray ``token slot'' represents the final marking. This net is able to replay all and only the sequences of activities that might have happened in reality.}
			\label{fig:behnet}
		\end{minipage}
	\end{figure}

	\subsection{Algorithms}\label{sec:algorithms}
	The algorithms contained in the PROVED tool are categorized in the three subpackages:
	\begin{itemize}
		\item \texttt{proved.algorithms.conformance}: this subpackage contains all the functionalities related to measuring conformance between uncertain data and a normative Petri net employing the alignment technique~\cite{pegoraro2019mining, pegoraro2020conformance}. It includes functions to compute upper and lower bounds for conformance score through exhaustive alignment of the realizations of an uncertain trace, and an optimized technique to efficiently compute the lower bound.
		\item \texttt{proved.algorithms.discovery}: this subpackage contains the functionalities needed to perform process discovery over uncertain event logs. It offers functionalities to compute a UDFG, a graph representing an extension of the concept of directly-follows relationship on uncertain data; this construct can be utilized to perform inductive mining~\cite{pegoraro2019discovering}.
		\item \texttt{proved.algorithms.simulation}: this subpackage contains some utility functions to simulate uncertainty within an existing event log. It is possible to add separately the different kinds of uncertainty described in the taxonomy of~\cite{pegoraro2020conformance}, while fine-tuning the dictionary of activity labels to sample and the amplitude of time intervals for timestamps.
	\end{itemize}

	\subsection{Interface}\label{sec:interface}
	Some of the functionalities of the PROVED tool are also supported by a graphical user interface. The PROVED interface is web-based, utilizing the Django framework in Python for the back-end, and the Bootstrap framework in Javascript and HTML for the front end. The user interface includes the PROVED library as a dependency, and is, thus, completely decoupled from the logic and algorithms in it. We will illustrate some parts of the user interface in the next section.

\section{Usage}\label{sec:usage}

In this section, we will outline how to install and use our tool. Firstly, let us focus on the programmatic usage of the Python library.

The full source code for PROVED can be found on the GitHub project page\footnote{Available at \url{https://github.com/proved-py/proved-core/}}. Once installed Python on the system, PROVED is available through the \texttt{pip} package manager for Python, and can be installed with the terminal command \texttt{pip install proved}, which will also install all the necessary dependencies.

Thanks to the import and export functionalities inherited from PM4Py, which has full XES~\cite{verbeek2010xes} certification, it is possible to start uncertain logs analysis easily and compactly. Let us examine the following example:

\footnotesize
\begin{Verbatim}[numbers=left, frame=single]
from pm4py.objects.log.importer.xes import importer as x_importer
from proved.artifacts import behavior_graph, behavior_net

uncertain_log = x_importer.apply('uncertain_event_log.xes')
uncertain_trace = uncertain_log[0]
beh_graph = behavior_graph.BehaviorGraph(uncertain_trace)
beh_net = behavior_net.BehaviorNet(beh_graph)
\end{Verbatim}
\normalsize

In this code snippet, an uncertain event log is imported, then the first trace of the log is selected, and the behavior graph and behavior net of the trace are obtained. Nodes and connections of behavior graphs and nets can be explored using the igraph functionalities and the PM4Py functionalities. We can also visualize both objects with Graphviz, obtaining graphics akin to the ones in Figures~\ref{fig:behgraph} and~\ref{fig:behnet}.

\footnotesize
\begin{Verbatim}[numbers=left, frame=single]
from pm4py.objects.petri.importer import importer as p_importer
from proved.algorithms.conformance.alignments import alignment_bounds_su

net, i_mark, f_mark = p_importer.apply('model.pnml')

alignments = alignment_bounds_su_log(uncertain_log, net, i_mark, f_mark)
\end{Verbatim}
\normalsize

In the snippet given above, we can see the code that allows to compute upper and lower bounds for conformance score of all the traces in the uncertain log against a reference model that we import, utilizing the technique of alignments~\cite{pegoraro2020conformance}. For each trace in the log, a pair of alignment objects is computed: the first one corresponds to an alignment with a cost equal to the lower bound for conformance cost, while the second object is an alignment with the maximum possible conformance cost. The object \texttt{alignments} is a list with one of such pairs for each trace in the log.

Let us now see some visual examples of the usages of the PROVED tool user interface\footnote{Available at \url{https://github.com/proved-py/proved-app/}}. The graphical tool can be executed in a local environment by starting the Django server in a terminal with the command \texttt{python manage.py runserver}.

\begin{figure}
	\centering
	\includegraphics[width=.8\textwidth, keepaspectratio]{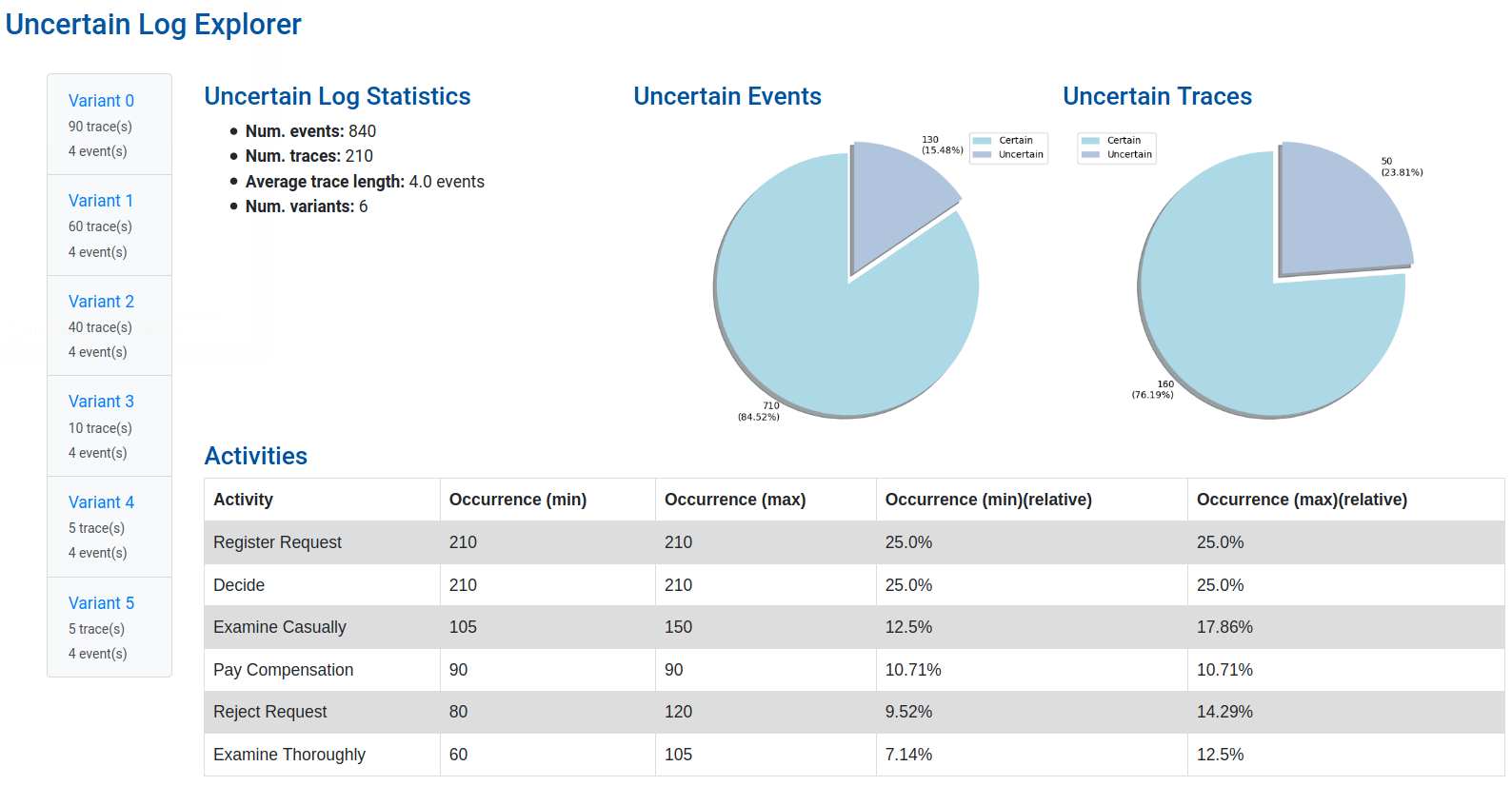}
	\caption{The dashboard of the PROVED user interface. This screen contains general information regarding an uncertain event log, including the list of uncertain variants, the number of instances of each activity label (minimum and maximum), and statistics regarding the frequency of uncertain events and uncertain traces in the log.}
	\label{fig:dashboard}
\end{figure}

Upon opening the tool and loading an uncertain event log, we are presented with a dashboard that summarizes the main information regarding the event log, as shown in Figure~\ref{fig:dashboard}.

In the center panel of the dashboard, we can see statistics regarding the uncertain log. On the top left, we find basic statistics such as the size of the log in the number of events and traces, the average trace length, and the number of uncertain variants. Note that the classical definition of \emph{variant} is inconsistent in uncertain event logs; rather, uncertain variants group together traces which have mutually isomorphic behavior graphs~\cite{pegoraro2020conformance}. We can also find pie charts indicating the percentage of uncertain events in the log (events with at least one uncertain attribute) and the percentage of uncertain traces in the log (traces with at least one uncertain event).

On the bottom, a table reports the counts of the number of occurrences for each activity label in the event log. Because of uncertainty on activity labels and indeterminate events, there is a minimum and maximum amount of occurrences of a specific activity label. The table reports both figures. There are two other tables in the dashboard, the Start Activities table and the End Activities table. Both are akin to the activity table depicted, but separately list activity labels appearing in the first or last event in a trace.

\begin{figure}
	\centering
	\includegraphics[width=.8\textwidth, keepaspectratio]{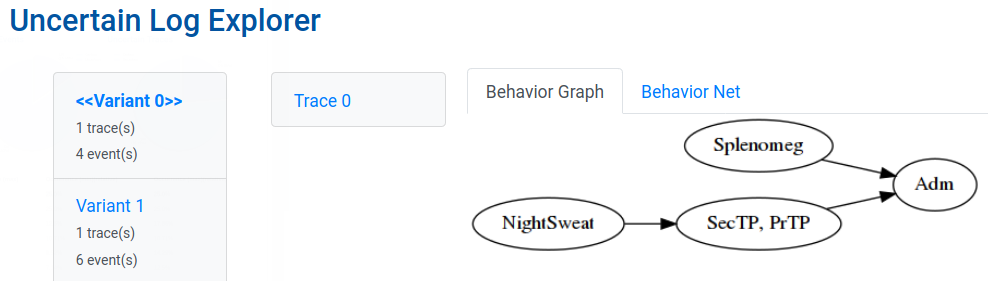}
	\caption{The uncertain variant page of the PROVED tool, showing information regarding the variant obtained from the trace in Table~\ref{table:uncertaintrace}. For a variant in an uncertain log, this page lists the traces belonging to that variant, and displays the graphical representations for that variant -- behavior graph and behavior net (the latter is not displayed, but can be accessed through the tab on the top).}
	\label{fig:bg}
\end{figure}

Upon clicking on one of the uncertain variants listed on the left, the user can access the graphical representation of the variant. It is possible to visualize both the behavior graph and the behavior net: the former is depicted in Figure~\ref{fig:bg}. The figure specifically shows information related to the trace depicted in Table~\ref{table:uncertaintrace}.

\begin{figure}
	\centering
	\includegraphics[width=.8\textwidth, keepaspectratio]{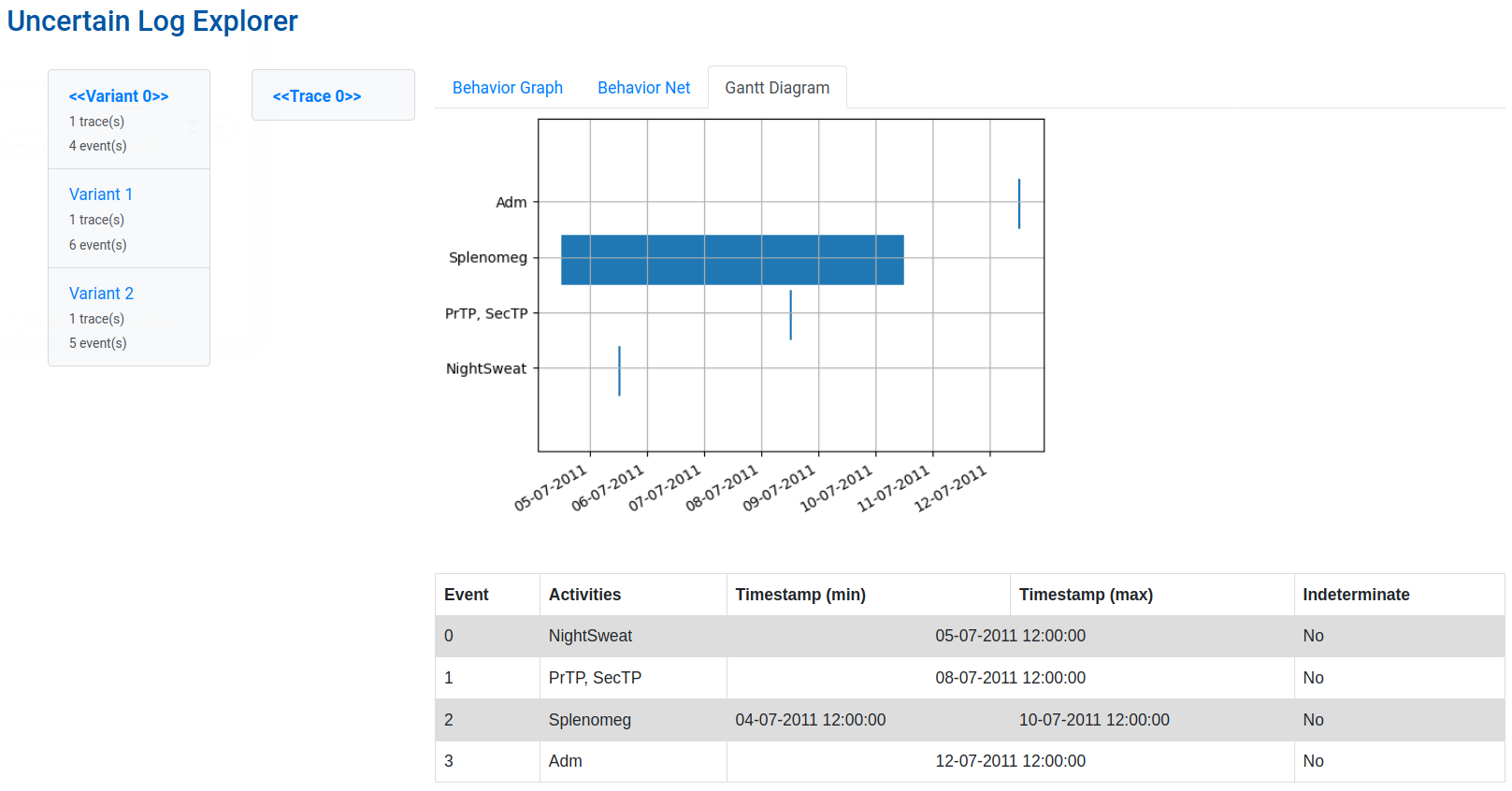}
	\caption{Visualization dedicated to a specific trace in the PROVED tool, showing information related to the trace in Table~\ref{table:uncertaintrace}. It is possible to see details on each event and on the uncertainty that might affect them, as well as a visualization showing the time relationship between uncertain event in scale.}
	\label{fig:gantt}
\end{figure}

Next to the variant menu on the left, we now have a trace menu, listing all the traces belonging to that uncertain variant. Clicking on a specific trace, the user is presented with data related to it, including a tabular view of the trace similar to that of Table~\ref{table:uncertaintrace}, and a Gantt diagram representation of the trace. Similarly to the behavior graph, the Gantt diagram shows time information in a graphical manner; but, instead of showing the precedence relationship between events, it shows the time information in scale, representing the time intervals on an absolute scale. This visualization is presented in Figure~\ref{fig:gantt}.

The interface allows the user to explore the features of an uncertain log, to ``drill down'' to variants, traces, event and single attributes, and visualize the uncertain data in a graphical manner without the need to resort to coding in Python.

Lastly, the menu on the left also allows for loading a Petri net, and obtaining alignments on uncertain event data.

\begin{figure}
	\centering
	\includegraphics[width=.8\textwidth, keepaspectratio]{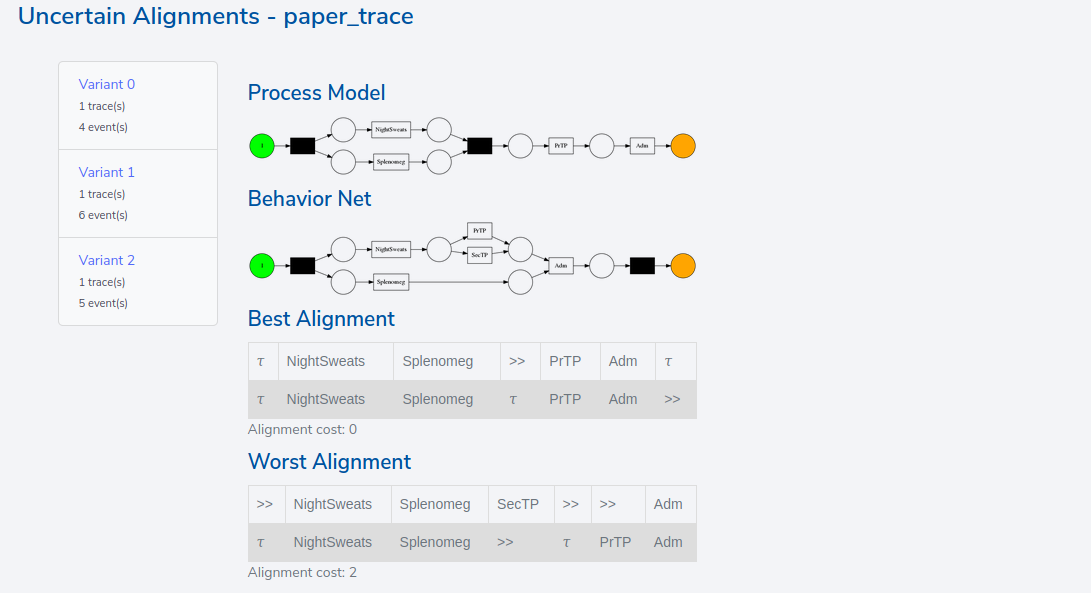}
	\caption{Visualization of alignments of the uncertain trace in Table~\ref{table:uncertaintrace} and a normative process model. In this case, the optimal alignment in the best case scenario perfectly fits the model, while in the worst case scenario we have an alignment cost of 2, caused by one move on model and one move on log.}
	\label{fig:alignments}
\end{figure}

As shown above, every uncertain trace can be represented by a behavior net. A conformance score can be computed between such behavior nets and a normative process model also represented by a Petri net: Figure~\ref{fig:alignments} illustrate the results of such alignment. For a given behavior net, two alignments are provided, together with the respective cost: one, showing a best-case scenario, and the other showing a worst-case scenario. This enables diagnostics on uncertain event data.

\section{Conclusions}\label{sec:conclusions}

In many real-world scenarios, the applicability of process mining techniques is severely limited by data quality problems. In some situations, these anomalies causing an erroneous recording of data in an information system can be translated in uncertainty, which is described through meta-attributes included in the log itself. Such uncertain event log can still be analyzed and mined, thanks to specialized process mining techniques. The PROVED tool is a Python-based software that enables such analysis. It provides capabilities for importing and exporting uncertain event data in the XES format, for obtaining graphical representations of data that can capture the behavior generated by uncertain attributes, and for computing upper and lower bounds for conformance between uncertain process traces and a normative model in the form of a Petri net.

Future work on the tool includes the definition of a formal XES language extension with dedicated tags for uncertainty meta-attributes, the further development of front-end functionalities to include more process mining capabilities, and more interactive objects in the user interface. Moreover, the research effort on uncertainty affecting the data perspective of processes can be integrated with the model perspective, blending uncertainty research with formalisms such as stochastic Petri nets.

\bibliographystyle{splncs04}
\bibliography{bibliography}

\end{document}